\documentclass{article} 
\usepackage{LIMARLVR,times}

\usepackage{wrapfig}
\usepackage{lipsum} 
\usepackage{wrapfig}
\usepackage[T1]{fontenc}

\usepackage{amsmath,amsfonts,bm}
\usepackage{multicol}

\def\eqref#1{equation~\ref{#1}}

\def\1{\bm{1}}

\DeclareMathAlphabet{\mathsfit}{\encodingdefault}{\sfdefault}{m}{sl}
\SetMathAlphabet{\mathsfit}{bold}{\encodingdefault}{\sfdefault}{bx}{n}

\usepackage{booktabs}      
\usepackage{multirow}      
\usepackage{amssymb}       
\usepackage{graphicx}      

\usepackage{listings}
\usepackage{courier}
\usepackage{graphicx}
\usepackage{subcaption} 
\usepackage{colortbl}
\usepackage{xcolor}

\usepackage{hyperref}
\usepackage{url}
\usepackage{multicol}
\usepackage{aliascnt}
\usepackage{adjustbox}
\usepackage{tcolorbox}
\usepackage{siunitx} 
\usepackage{booktabs}
\usepackage{multirow} 
\usepackage{array}
\usepackage{enumitem}
\usepackage{booktabs} 
\usepackage{amssymb}  
\usepackage{amsmath}  
\usepackage{graphicx} 
\usepackage{longtable} 
\usepackage{CJKutf8} 
\definecolor{uclablue}{rgb}{0.15, 0.45, 0.68}
\hypersetup{
    breaklinks,
    citecolor=uclablue,
    colorlinks=true,
}

\newcommand{\hi}[1]{\textbf{\textcolor{teal!60!blue}{#1}}}

\title{DRIVE: \texorpdfstring{\hi{D}ata Curation Best Practices for \hi{R}einforcement Learning w\hi{I}th \hi{V}\hi{E}rifiable Reward in Competitive Code Generation}{DRIVE: Data Curation Best Practices for Reinforcement Learning with Verifiable Reward in Competitive Code Generation}}

\author{\textbf{Speed Zhu, Jianwei Cai, Guang Chen, Lulu Wu, Saiyong Yang, Wiggin Zhou}%
\thanks{Corresponding author: wigginzhou@tencent.com}%
\thanks{Emails: speedzhu@tencent.com; cmathxcai@tencent.com; maxluxchen@tencent.com; lukywu@tencent.com; stevesyang@tencent.com}
\\[1.5ex] 
  Hunyuan Team, Tencent
}

\begin{document}

\maketitle\let\oldthefootnote\thefootnote

\begin{abstract}
Recent reasoning-first models (e.g., OpenAI o1, DeepSeek R1) have spurred a resurgence of interest in RLVR. Nevertheless, advances are dominated by mathematics (e.g., AIME), with competitive-programming code generation underexplored and data curation receiving less attention than RL algorithm design. We investigate how to construct RLVR datasets (i.e., RL prompts) and present practical training techniques that yield strong performance on competitive-programming code generation. Our pipeline begins with supervised fine-tuning (SFT) distilled from strong open-source models, augmented with general-purpose and reasoning-intensive data. RL then follows a two-stage process with executable, testcase-driven rewards: first, training on a large, uniformly distributed set of competitive-programming problems using Group Relative Policy Optimization (GRPO) with 8 rollouts per prompt and a relatively short response-generation window (e.g., 32k during SFT and 24k in this stage) to expand entropy and mitigate repetition and truncation; second, we perform \textbf{Pre-GRPO}: updating on a small, high-quality set of challenging problems with a large rollout budget (64 rollouts per prompt) under a hard-focus curriculum that continuously retains the most difficult instances throughout training. We implement our method on Qwen2.5-32B and evaluate on LeetCode and Codeforces weekly contests to avoid data leakage. The resulting model achieves state-of-the-art performance among models of similar scale and is comparable to leading systems such as DeepSeek v3.1 and Doubao-1.5-Thinking. We also examine scaling trends and observe strong RL scaling on an internal large-scale MoE model. Our study distills concise best practices for data curation, entropy expansion, and curriculum design in RLVR for competitive-programming code generation.
\end{abstract}


\begin{figure}[!ht]
    \centering
    \includegraphics[width=0.9\textwidth]{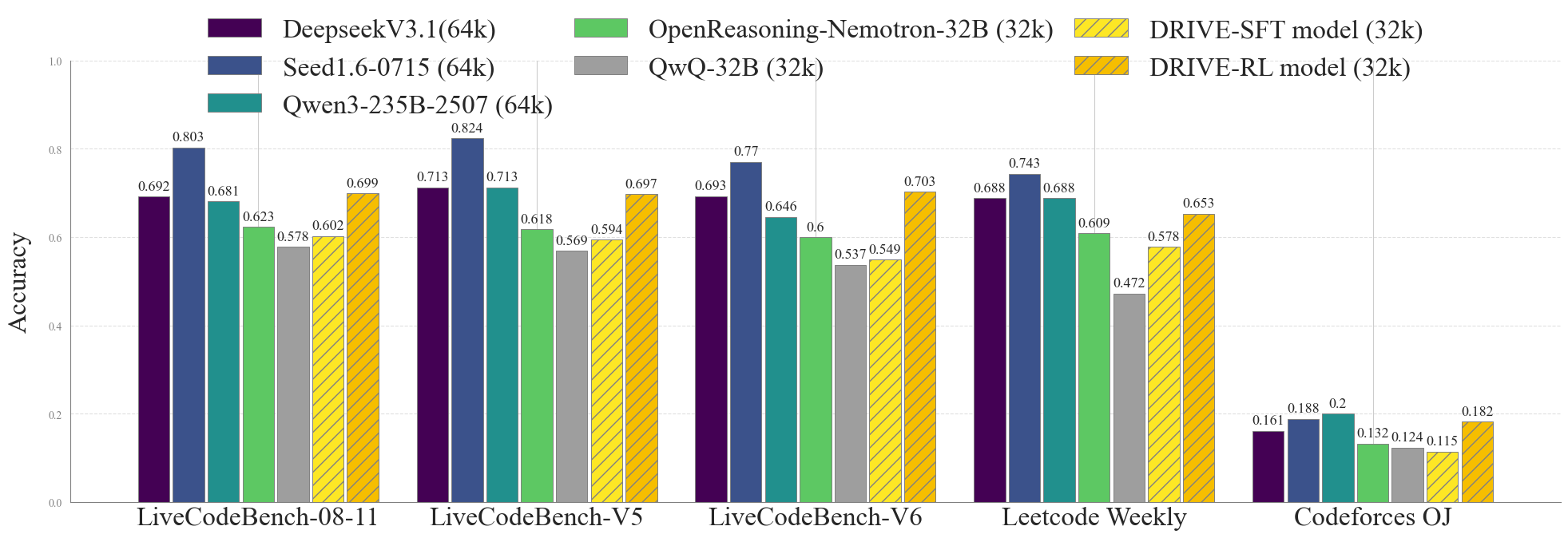}
    \caption{Performance of our models on various benchmark}
    \label{fig:performance}
\end{figure}

\section{Introduction}
\label{sec:intro}

Recent advances in reasoning-first language models, exemplified by OpenAI o1~\citep{jaech2024openai} and DeepSeek R1~\citep{guo2025deepseek}, have demonstrated remarkable capabilities in complex problem-solving tasks through reinforcement learning with verifiable rewards (RLVR). Although a large body of research centers on math benchmarks such as AIME~\citep{yu2025dapo,li2025treepo, lou2025adacot, yue2025vapo, liu2025prorl}, applying these models to competitive programming—a domain demanding both deep algorithmic insight and precise implementation—remains relatively underexplored~\citep{zhao2024marco, zhang2024policy}. This gap is particularly notable given that competitive programming presents unique challenges: solutions must not only be logically correct but also executable, efficient, and capable of handling edge cases within strict computational constraints. 

Additionally, prior work on RLVR has predominantly focused on developing novel algorithms and training techniques~\citep{shao2024deepseekmath, hu2025reinforce++, zheng2025group, li2025knapsack}, with considerably less attention paid to the critical aspects of data curation and curriculum design~\citep{shen2025exploring}. Most existing approaches either apply uniform training strategies across all problem difficulties or rely on simplistic difficulty categorization~\citep{seed2025seed1, yu2025dapo, li2025treepo, li2023remax}, potentially limiting the model's ability to tackle the most challenging problems that define competitive programming excellence. Furthermore, the computational demands of training large-scale models often make extensive experimentation prohibitive, necessitating more efficient and targeted training strategies.

In this work, we present a comprehensive study on applying RLVR to competitive programming code generation, with particular emphasis on practical data curation techniques and curriculum design. Our two-stage RL framework directly addresses limitations in standard SFT and RLVR, increasing exploration entropy, reducing repetitive generation, and improving performance on challenging problems. In stage one, we expand exploration entropy with a large, uniformly distributed pool of competitive-programming problems and moderate rollout budgets. In stage two, Pre-GRPO, we adopt a hard-focus curriculum and a high rollout budget to master the most challenging cases.

We implement our method on Qwen2.5-32B~\citep{hui2024qwen2} and demonstrate its effectiveness through extensive evaluation on recent LeetCode and Codeforces weekly contests, benchmarks carefully selected to avoid data contamination. Our model achieves state-of-the-art performance among similarly sized models and remains competitive against much larger model such as DeepSeek V3.1~\citep{tacsyurek2025comparative}, with particularly strong gains on challenging problems (up to 58\% relative improvement over similarly sized models on Codeforces). Our comprehensive ablation studies confirm that entropy expansion delivers robust generalization and the hard-focus curriculum extends the model’s problem-solving frontier. In addition, we observe strong RL scaling on an internal large-scale MoE model.

Our key contributions are:
\begin{itemize}[left=0pt]
    \item \textbf{A two-stage RL framework} that systematically addresses the limitations of normal RLVR methods through entropy expansion followed by hard-focus curriculum learning.
    \item \textbf{Empirical evidence} demonstrating that large rollout budgets are crucial for learning challenging problems, while moderate budgets suffice for entropy expansion.
    \item \textbf{Comprehensive analysis} of training dynamics revealing that standard RL struggles with difficult cases, motivating our curriculum-based approach.
    \item \textbf{State-of-the-art results} on competitive programming benchmarks with a 32B parameter model, achieving performance comparable to models with 5--10$\times$ more parameters.
    \item \textbf{Practical insights} for RLVR implementation, including data curation strategies, difficulty-aware curriculum design, and scaling trends validated on an internal large-scale MoE model
\end{itemize}

Our findings suggest that careful curriculum design and strategic use of computational resources can yield substantial improvements in competitive programming capabilities, providing a roadmap for future work in this challenging domain.
\section{Related Work}
\textbf{RLVR Algorithms.} Since DeepSeek R1~\citep{guo2025deepseek} demonstrated successful RLVR implementation on both Qwen-32B and DeepSeekV3, numerous studies have investigated performance improvements, particularly on the AIME benchmark. DAPO~\citep{yu2025dapo} was the first open-source method to enhance RLVR performance on AIME using the GRPO algorithm with Qwen2.5-32B. Subsequently, VAPO~\citep{yue2025vapo} proposed an enhanced PPO algorithm for further gains, while ProRL~\citep{liu2025prorl} explored GRPO optimization tricks to boost benchmark performance. Other works have focused on stabilizing RLVR training, including expert replay strategies for MoE models~\citep{zheng2025group} and truncated importance sampling to address mismatches between inference and training engines~\citep{yao2025offpolicy}. However, despite these algorithmic advances, remarkably little attention has been paid to the critical aspect of constructing suitable RL prompts to enhance RLVR performance—a gap our work addresses.

\textbf{RLVR Data Construction.} Few studies examine how to construct RL training prompts to enhance RLVR or RLHF. \citet{gao2025principled} propose a principled data-selection method for DPO, showing that overly difficult examples hinder alignment and should be filtered during training. \citet{li2025limr} introduce a strategic selection procedure that identifies key prompts from a full set, achieving comparable RLHF performance with only a subset of the data. \citet{shen2025exploring} propose Pre-PPO, a data-selection algorithm that enables RL scaling in RLHF by selecting suitable examples from large datasets. However, RL data construction for RLVR, especially for competitive-programming code generation, remains largely unexplored. To our knowledge, this is the first study to systematically investigate this problem in RLVR, and we analyze both performance and scaling trends on larger models.

\textbf{RLVR performance scaling analysis.} Although recent works—e.g., DeepSeek R1~\citep{guo2025deepseek}, SEED-1.5-Thinking~\citep{seed2025seed1}, and Qwen3~\citep{yang2025qwen3}—present RLVR strategies across multiple benchmarks, few report performance on both small and large models; notable exceptions include DeepSeek R1 and Qwen3. In the RLHF domain, \citet{shen2025exploring} provide a comprehensive analysis of scaling trends. However, most RLVR evaluations focus on dense 7–32B models, leaving open questions about scaling to much larger models. In this paper, we conduct a comprehensive performance and strategy ablation on Qwen2.5-32B and then directly apply the resulting recipe to on an internal large-scale MoE model, achieving strong results. This demonstrates that our strategy is effective and suitable for scaling up.

\section{Method}
\label{sec:method}
\begin{figure}[!ht]
    \centering
    \includegraphics[width=0.9\textwidth]{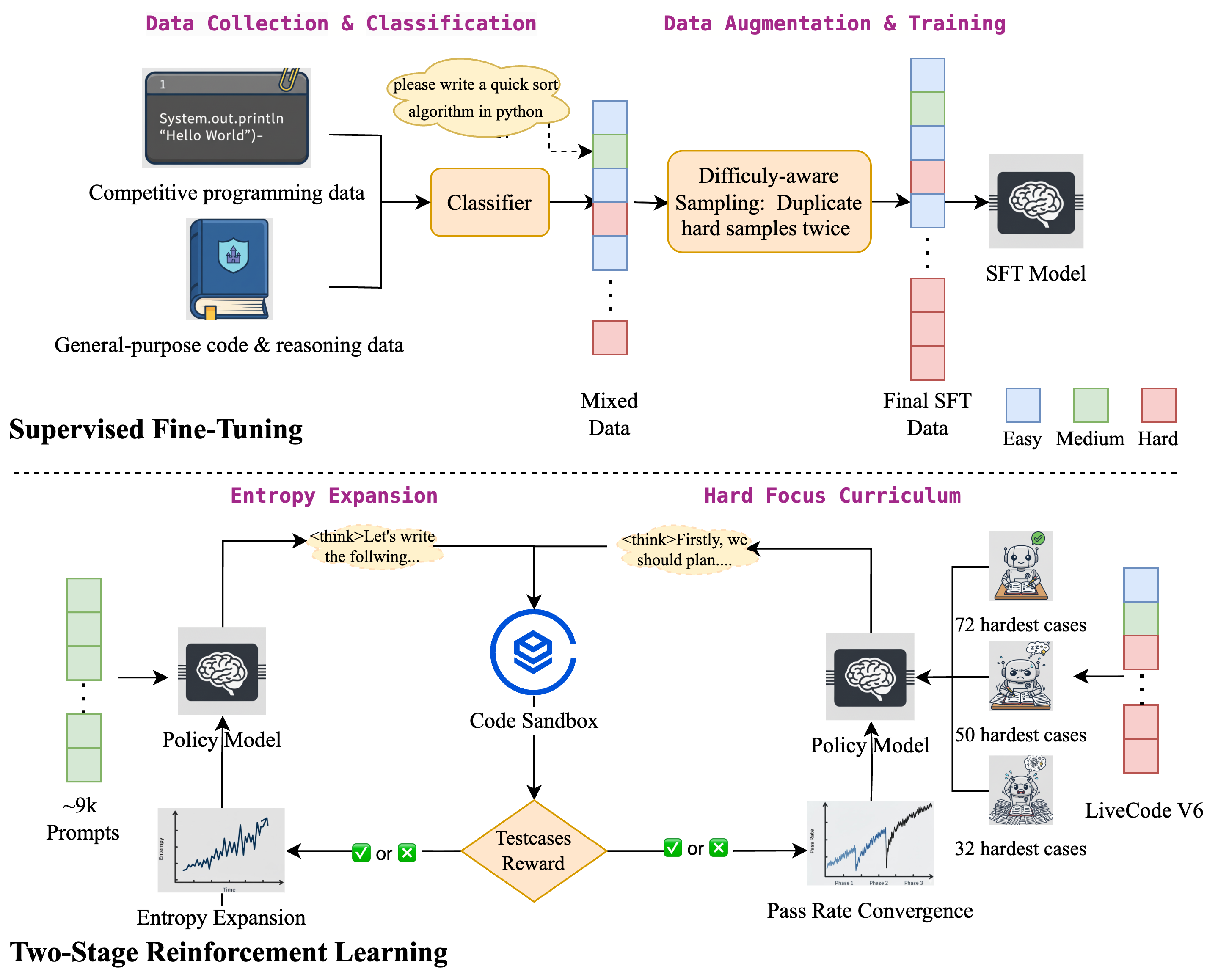}
    \caption{The training pipeline of our models.}
    \label{fig:main_process}
\end{figure}
Our training pipeline is illustrated in Figure \ref{fig:main_process}.
Our approach consists of two main phases: supervised fine-tuning followed by a two-stage reinforcement learning process with verifiable rewards.
\subsection{Supervised Fine-Tuning}
We begin with supervised fine-tuning (SFT) on Qwen2.5-32B, distilling knowledge from strong open-source models. The training data is augmented with general-purpose and reasoning-intensive datasets to enhance the model's problem-solving capabilities. 

Specifically, we first collect competitive programming prompts from open-source datasets. We then train a small model to classify problem difficulty and assign difficulty labels to each problem, categorizing them into three types: easy, medium, and hard. We duplicate hard problems twice in the training data, following a similar approach to Difficulty-Aware Rejection Tuning for Mathematical Problem-Solving~\citep{tong2407difficulty}. However, we do not directly adopt their RFT method, as it requires extensive model sampling which is computationally expensive for our use case. Our simpler duplication strategy achieves similar benefits of emphasizing difficult problems while maintaining computational efficiency. Furthermore, we collect additional prompts from other code generation tasks beyond competitive programming. We find that including both general-purpose coding data and reasoning-intensive problems significantly improves performance on competitive programming tasks, likely due to enhanced reasoning and code comprehension abilities.

\begin{figure}[!ht]
    \centering
    \includegraphics[width=0.7\textwidth]{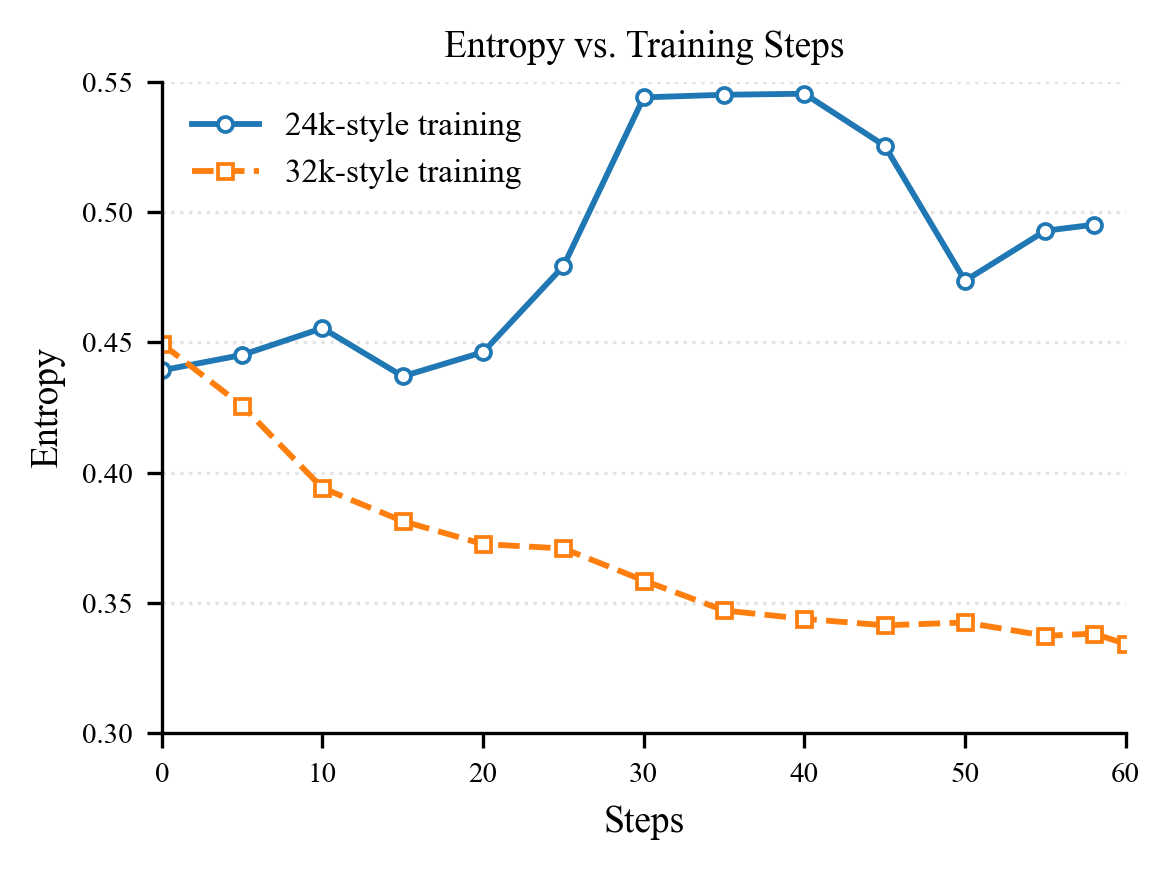}
    \caption{The entropy comparison of 24k-style training and 32k-style training}
    \label{fig:entropy}
\end{figure}

\begin{table}[t]
\centering
\caption{Results on LeetCode Weekly OJ (32 problems) and Codeforces OJ (33 problems).}
\label{tab:oj_results}
\begin{tabular}{lcccc}
\toprule
\multirow{2}{*}{\textbf{Method}} & \multicolumn{2}{c}{\textbf{LeetCode Weekly OJ (32)}} & \multicolumn{2}{c}{\textbf{Codeforces OJ (33)}} \\
\cmidrule(lr){2-3} \cmidrule(lr){4-5}
 & Pass@10 & Avg@1 & Pass@10 & Avg@1 \\
\midrule
SFT Model          & \colorbox{green!20}{96.88\%} & 57.81\% & \colorbox{red!20}{24.24\%} & 11.52\% \\
DeepseekV3.1       & \colorbox{green!20}{96.88\%} & 68.75\% & \colorbox{orange!20}{33.33\%} & 16.06\% \\
Seed1.6-0715       & \colorbox{green!20}{96.88\%} & 74.38\% & \colorbox{orange!20}{\textbf{39.39\%}} & 18.79\% \\
\bottomrule
\end{tabular}
\end{table}

\subsection{Two-Stage Reinforcement Learning}
\textbf{Analysis of the SFT Model for the RL Phase.} Following extensive distilled SFT, we identify three critical issues that motivate our two-stage RL design. First, as shown in Figure~\ref{fig:entropy}, the model exhibits \textbf{low entropy} and limited exploration, repeatedly converging to similar solution modes. Second, it shows \textbf{repetitive generation patterns}, often producing redundant code structures or falling into loops that lead to truncated outputs (see Sec.~\ref{app:case} for details). Third, \textbf{harder problems tend to require longer responses}, which are harder to learn and generalize during SFT due to long-sequence dependency challenges. Empirically (Table~\ref{tab:oj_results}), the SFT model’s pass@10 on the LeetCode Weekly OJ is comparable to DeepSeek V3.1 and Seed1.6-0715, but on the (harder) Codeforces Weekly OJ its pass@10 lags behind both. These observations directly inform our RL strategy: Stage~1 targets entropy expansion and pattern diversification, while Stage~2 strengthens performance on challenging problems.

\textbf{Stage 1: Entropy Expansion.} In the first stage, we perform training on a mixed dataset of approximately 9k competitive programming problems. Each prompt is sampled with 8 rollouts with a total sequence length of 24k tokens (including both prompt and response) to increase output entropy and mitigate mode collapse. This stage serves several critical purposes: (1) \textit{Entropy Enhancement} - training on uniformly distributed, medium-difficulty problems expands the model's output diversity; (2) \textit{Pattern Reduction} - the increased entropy reduces repetitive generation patterns, significantly decreasing truncated errors; (3) \textit{Overall Performance} - this training stage provides substantial improvements to the model's general competitive programming capabilities.

\textbf{Stage 2: Hard-Focus Curriculum.} The second stage focuses on high-quality, challenging problems using LiveCode V6 data filtered by Pre-GRPO. Pre-GRPO carries forward a subset of low–pass-rate cases from each phase to the next. Specifically, we train for 32k steps under a three-phase progressive curriculum: Phase 1 uses the 72 hardest cases with a 64-step budget, Phase 2 the 50 hardest with a 32-step budget, and Phase 3 the 25 hardest with a 32-step budget. We sample 80 rollouts per prompt in this stage, as large rollout sampling is crucial for stable performance gains. This curriculum continuously retains the most difficult instances, pushing the model to master increasingly challenging problems while maintaining performance on easier cases.
\section{Experiments}
\subsection{Experimental Setup}
We describe our experimental setup as follows:
\begin{itemize}[left=0pt]
    \item \textbf{Models:} We conducted experiments with the open-source Qwen2.5-32B-Instruct and trained our SFT model on top of this foundation. We also examined scaling trends on an internal large-scale MoE model.
    
    \item \textbf{Data:} For SFT data, we initially collected 1.27M open-source prompts and generated corresponding responses through distillation from the open-source model DeepSeekR1-0528. We then refined this dataset to 470K high-quality prompts using a 5-round arena learning method. For the first stage of RL training, we utilized 9K prompts from open-source repositories. In the second stage of RL, we employed the LiveCode V6 dataset, which comprises 175 high-quality examples with comprehensive test cases.

    \item \textbf{Experimental Details of SFT:} We trained the model on the selected 470K SFT data for 3 epochs using the Qwen2.5-32B-Instruct as our base model. The training was conducted with a learning rate of $1 \times 10^{-5}$, utilizing 256 GPUs with a global batch size of 512.

    \item \textbf{Experimental Details of RL:} In the \textit{Entropy Expansion RL stage}, we trained the model on 9K prompts for 32 steps by GRPO algorithm. Each prompt is sampled with 8 rollouts with a total sequence length of 24k tokens. In the \textit{Hard-Focus Curriculum RL} stage, we continued training on progressively harder problems as specified in Section~\ref{sec:method}, where each prompt is sampled with 64 rollouts with a total sequence length of 32k tokens.  

    \item \textbf{Evaluations:} We construct a comprehensive evaluation set covering competitive code generation tasks. Specifically, we adopt LiveCode08-11 (166 problems), LiveCodeV5 (167 problems), and LiveCodeV6 (175 problems) benchmarks as our validation sets. While LiveCodeV6 was exposed during the second RL stage, both LiveCode08-11 and LiveCodeV5 remain uncontaminated throughout the SFT and RL stages. Furthermore, to ensure the integrity of our experimental results and avoid data leakage, we evaluate our models on recent LeetCode and Codeforces weekly contests that were released after our training data collection cutoff.
\end{itemize}

\begin{table}[t]
\centering
\caption{Pass@1 performance comparison across different evaluation benchmarks.}
\label{tab:benchmark_comparison}
\begin{tabular}{lccccc}
\toprule
\textbf{Model} & \textbf{LiveCode} & \textbf{LiveCode} & \textbf{LiveCode} & \textbf{LeetCode} & \textbf{Codeforces} \\
 & \textbf{08-11} & \textbf{V5} & \textbf{V6} & \textbf{Weekly (32)} & \textbf{OJ (33)} \\
\midrule
DeepseekV3.1 (64k) & 0.692 & 0.713 & 0.693 & 0.688 & 0.161 \\
\rowcolor{red!10}
Seed1.6-0715 (64k) & \textbf{0.803} & \textbf{0.824} & \textbf{0.770} & \textbf{0.743} & 0.188 \\
Qwen3-235B-2507 (64k) & 0.681 & 0.713 & 0.646 & 0.688 & \textbf{0.200} \\
QwQ-32B (32k) & 0.578 & 0.569 & 0.537 & 0.472 & 0.124 \\
OpenReasoning-Nemotron-32B (32k) & 0.623 & 0.618 & 0.600 & 0.609 & 0.132 \\
\midrule
SFT model (32k) & 0.602 & 0.594 & 0.549 & 0.578 & 0.115 \\
RL Stage 1 model (24k) & 0.625 & 0.627 & 0.634 & 0.603 & 0.112 \\
\rowcolor{blue!10}
RL model (32k) & \textbf{0.699} & \textbf{0.697} & \textbf{0.703} & \textbf{0.653} & \textbf{0.182} \\
\midrule
\rowcolor{green!20}
\textbf{Relative Improvement (RL vs SFT)} & \textcolor{green!70!black}{\textbf{+16.1\%}} & \textcolor{green!70!black}{\textbf{+17.3\%}} & \textcolor{green!70!black}{\textbf{+28.1\%}} & \textcolor{green!70!black}{\textbf{+13.0\%}} & \textcolor{green!70!black}{\textbf{+58.3\%}} \\
\bottomrule
\end{tabular}
\end{table}

\subsection{Experimental Results}
The experimental results presented in Table 1 demonstrate the
following key findings:
\begin{itemize}[left=0pt]
\item \textbf{Competitive Performance with Smaller Scale}: Despite having only 32B parameters and 32k context length, our RL model achieves comparable or superior performance to mainstream large-scale models like DeepSeek-V3.1 (64k), with particularly strong results on LiveCode 08-11 and V5 benchmarks (0.699-0.697 vs 0.692-0.713), demonstrating remarkable efficiency in the parameter-performance trade-off.

\item \textbf{Significant Advancement Over 32B SOTA}: Our RL model substantially outperforms the 32B state-of-the-art OpenReasoning-Nemotron-32B across all benchmarks, with improvements ranging from +0.076 (08-11) to +0.079 (V6) on LiveCode benchmarks and a notable +0.056 improvement on the challenging Codeforces OJ (0.182 vs 0.132), representing a 37.8\% relative improvement on competitive programming tasks.

\item \textbf{Substantial Improvement Over SFT Baseline}: The RL training delivers consistent and significant improvements across all evaluation benchmarks, with absolute gains ranging from +0.077 (08-11) to +0.103 (V6), representing relative improvements of 16.1\% (08-11) to 17.3\% (V6) on LiveCode benchmarks.

\item \textbf{Robust Generalization Across Diverse Benchmarks}: Our model demonstrates strong generalization capabilities, achieving substantial improvements on both easier benchmarks (LiveCode series) and challenging competitive programming tasks (Codeforces), indicating effective learning of fundamental coding principles.

\item \textbf{Effective RL Training Strategy}: The stepwise improvement from SFT to RL Stage 1 to the final RL model validates our approach. The RL model delivers the largest gains on complex reasoning tasks, with 13.0\% relative improvement on LiveCode weekly OJ and 58.3\% on Codeforces weekly OJ.
\end{itemize}

\subsection{SFT Experiments}
\textbf{Ablation Study.} To assess how different SFT strategies affect model generalization, we conduct an ablation comparing three methods. Results are shown in Table~\ref{tab:ablation_diff_sft}.
\begin{itemize}[left=0pt]
\item \textbf{Basic SFT strategy:} The baseline model is trained on the full dataset of 1.27M prompts.
\item \textbf{Arena Learning strategy~\citep{luo2024arena}:} An iterative curriculum-learning approach. The dataset is split into five folds. We train an initial model on the first fold, use it to make predictions on the next fold, and retain only the samples the model fails on (``hard samples'') for the subsequent training stage. Repeating this process yields a condensed dataset of 470K hard samples.
\item \textbf{Twice Hard Learning strategy:} Building on Arena Learning, this method first identifies the hard samples and then doubles their exposure during training, keeping the overall training token budget comparable to the Basic SFT strategy while focusing computation on more challenging data.
\end{itemize}

As shown in the Table \ref{tab:ablation_diff_sft},  the \textbf{Twice Hard Learning Strategy} almostly outperforms the other methods across all evaluated benchmarks, achieving the best results.

\textbf{Key Findings.} \textit{Allocating a larger training token budget to hard samples proves effective.} We observe that the Arena Learning Strategy, despite reducing the training dataset from 1.27M to 470K prompts (a reduction of over 60\%), maintains performance comparable to the Basic SFT Strategy. This indicates that focusing on hard samples is an efficient approach. However, its slight underperformance can be attributed to an insufficient training token budget, which leads to incomplete learning during the SFT stage. To validate this hypothesis, the Twice Hard Learning Strategy was designed to restore the token budget by specifically oversampling the identified hard examples. The superior performance of this strategy, as shown in Table \ref{tab:ablation_diff_sft}, confirms our hypothesis. This demonstrates that for optimal performance, it is crucial not only to identify challenging data but also to allocate sufficient computational resources for the model to learn from them effectively.

\begin{table}[t]
\centering
\newcolumntype{L}[1]{>{\raggedright\arraybackslash}p{#1}}

\caption{Ablation study on different SFT training strategies.}
\label{tab:ablation_diff_sft}
\begin{tabular}{L{4.8cm} ccccc} 
\toprule
\textbf{Method} & 
\begin{tabular}{@{}c@{}}LiveCode \\ \textbf{08-11}\end{tabular} & 
\begin{tabular}{@{}c@{}}LiveCode \\ \textbf{V5}\end{tabular} & 
\begin{tabular}{@{}c@{}}LiveCode \\ \textbf{V6}\end{tabular} & 
\begin{tabular}{@{}c@{}}LeetCode \\ \textbf{Weekly (32)}\end{tabular} & 
\begin{tabular}{@{}c@{}}Codeforces \\ \textbf{OJ (33)}\end{tabular} \\
\midrule
Basic SFT Strategy & 0.582 & \textbf{0.603}&0.545& 0.558 & 0.112 \\
Arena Learning Strategy & 0.600& 0.598 & 0.542 & 0.553 &0.111 \\
\rowcolor{green!20}
\textbf{Twice Hard Learning Strategy} & \textbf{0.602}&0.594&\textbf{0.549}&\textbf{0.578}&\textbf{0.115} \\
\bottomrule
\end{tabular}
\end{table}

\begin{table}[t]
\centering
\newcolumntype{L}[1]{>{\raggedright\arraybackslash}p{#1}}

\caption{Ablation study on different RL training strategies.}
\label{tab:ablation_study_wrapped}
\begin{tabular}{L{4.8cm} ccccc} 
\toprule
\textbf{Method} & 
\begin{tabular}{@{}c@{}}LiveCode \\ \textbf{08-11}\end{tabular} & 
\begin{tabular}{@{}c@{}}LiveCode \\ \textbf{V5}\end{tabular} & 
\begin{tabular}{@{}c@{}}LiveCode \\ \textbf{V6}\end{tabular} & 
\begin{tabular}{@{}c@{}}LeetCode \\ \textbf{Weekly (32)}\end{tabular} & 
\begin{tabular}{@{}c@{}}Codeforces \\ \textbf{OJ (33)}\end{tabular} \\
\midrule
SFT Model & 0.602 & 0.594 & 0.549 & 0.578 & 0.115 \\
\rowcolor{red!15} 
RL with all LiveCodeV6 dataset & 0.506 & 0.512 & 0.522 & 0.296 & 0.105 \\
RL with all 9k data & 0.676 & 0.688 & 0.675 & 0.592 & 0.102 \\
RL without First Stage Only Second Stage & 0.636 & 0.626 & 0.691 & 0.550 & 0.142 \\
\rowcolor{green!20}
\textbf{Our Method} & \textbf{0.699} & \textbf{0.697} & \textbf{0.703} & \textbf{0.653} & \textbf{0.182} \\
\rowcolor{green!20}
\textbf{Our Method with More Harder Samples in Stage 2} & \textbf{0.712} & \textbf{0.707} & \textbf{0.743} & \textbf{0.678} & \textbf{0.188} \\
\bottomrule
\end{tabular}
\end{table}

\subsection{The RL Experiments}

\begin{figure}[ht]
  \centering
  \begin{subfigure}[t]{0.48\textwidth}
    \centering
    \includegraphics[width=\linewidth]{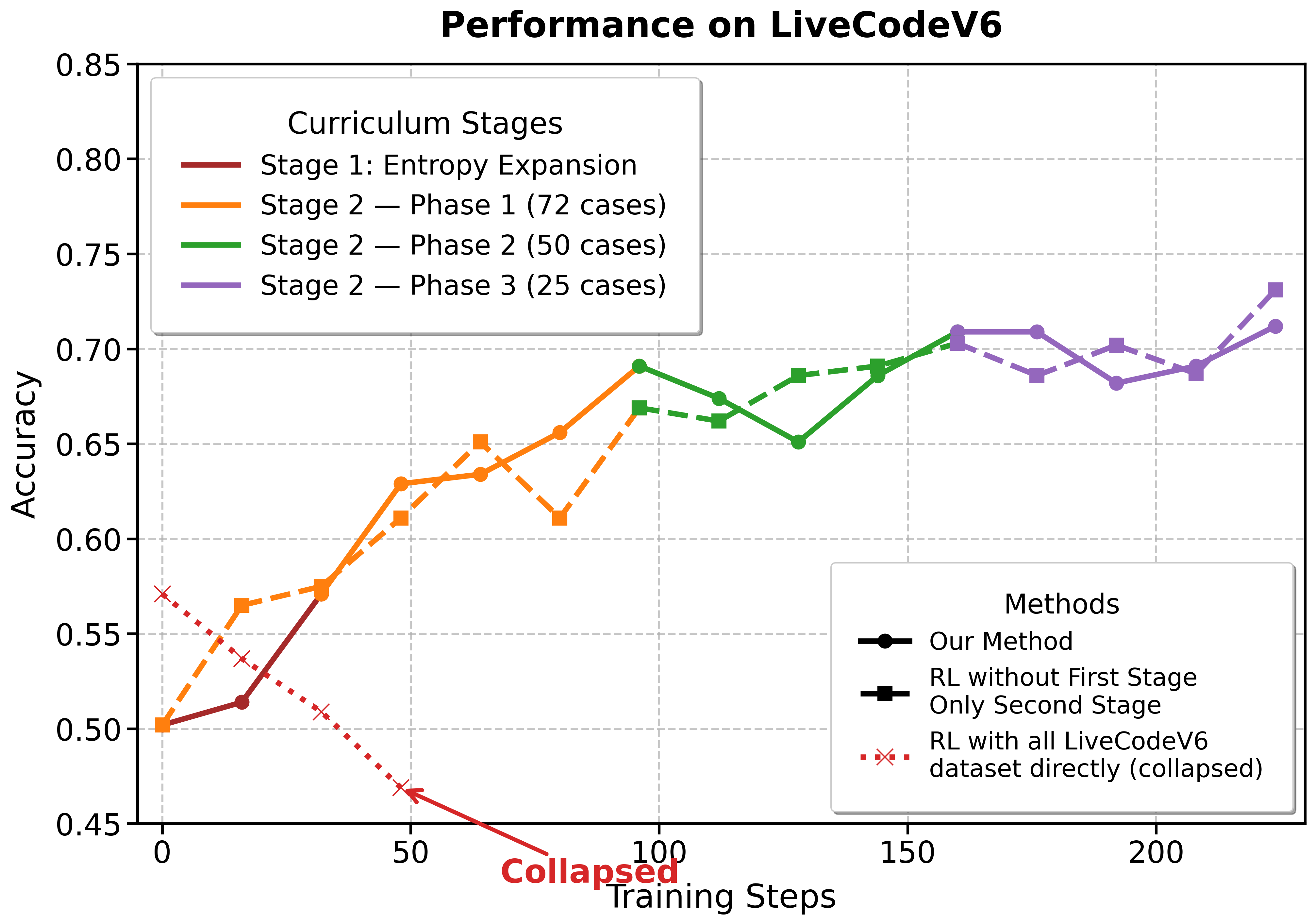}
    \caption{The performance on LiveCodeV6 during training}
    \label{fig:left}
  \end{subfigure}\hfill
  \begin{subfigure}[t]{0.48\textwidth}
    \centering
    \includegraphics[width=\linewidth]{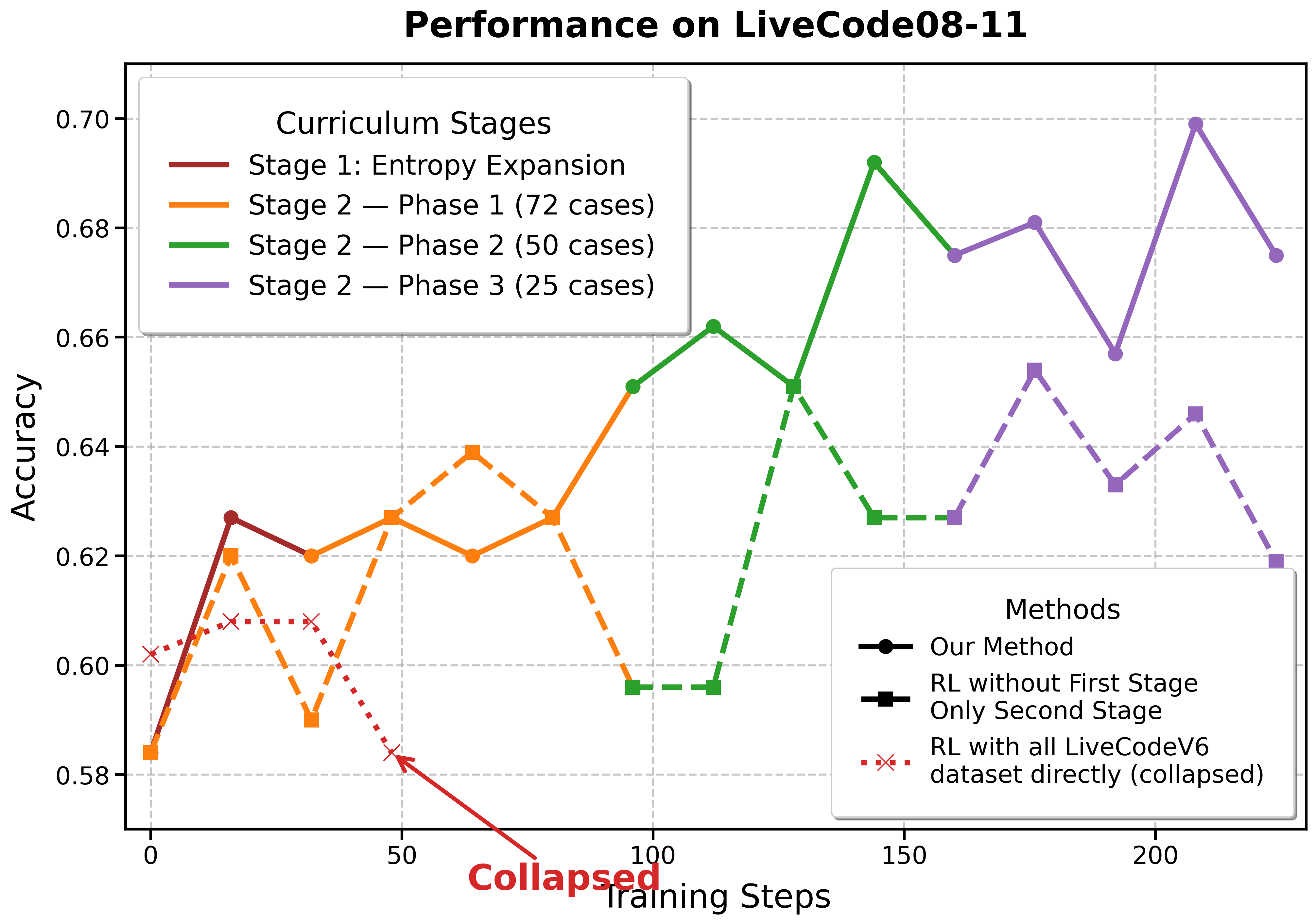}
    \caption{The performance on LiveCode08-11 during training}
    \label{fig:right}
  \end{subfigure}
  \caption{The performance of different RL training strategies on LiveCodeV6 and LiveCode08-11 during training}
  \label{fig:side-by-side}
\end{figure}

\textbf{Ablation Study.} To validate the effectiveness of our two-stage RL approach, we conduct ablation experiments by removing key components from our method. Table~\ref{tab:ablation_study_wrapped} and Figure~\ref{fig:side-by-side} present the results of three ablation variants compared to our full method, showing performance on both LiveCodeV6 (training set) and LiveCode08-11 (validation set) during the training process. Specifically, we examine: (1) \textit{RL with all LiveCode V6 dataset}, which applies 24k-style RLVR using all 175 LiveCodebench V6 problems directly on the SFT model; (2) \textit{RLVR with all 9k data}, which uses the complete 9k training dataset from the first stage to perform 32k-style RLVR on the SFT model; and (3) \textit{RL without First Stage (Second Stage Only)}, which directly applies the Stage 2 RLVR strategy (training only on the harder samples) to the SFT model, bypassing the first stage entirely.

\textbf{Key Findings:} \textit{Difficulty-aware training is crucial.} Difficulty-aware training is crucial. RL works best when driven by challenging samples: focusing Stage 2 RLVR on the hard subset yields stable improvements. In contrast, mixing easy problems into RL (training on the full 9k or the entire LiveCodeV6 without hardness filtering) fails to obtain best performance  and can even cause training/performance to collapse, with the largest drop on LeetCode Weekly (-48.8\%). This indicates that hard samples provide a useful reward signal, whereas easy samples dilute the signal and even destabilize optimization. (2) \textit{Entropy expansion stage matters.} Removing the first stage and only applying hard-focus curriculum RL (row 3) yields mixed results—while it improves on contaminated LiveCodeV6 (+25.9\%), it actually hurts performance on out-of-distribution benchmarks like LeetCode weekly OJ (-4.8\%). This suggests that the entropy expansion stage helps the model develop robust problem-solving capabilities that generalize beyond the training distribution. (3) \textit{Both stages synergize effectively.} Our complete two-stage method (highlighted in green) achieves the best performance across all benchmarks, with improvements ranging from 13.0\% on LeetCode weekly OJ to 58.3\% on Codeforces weekly OJ over the SFT baseline, confirming that the combination of entropy expansion followed by hard-focus curriculum learning is essential for optimal performance.

\begin{figure}[!ht]
    \centering
    \includegraphics[width=1.0\textwidth]{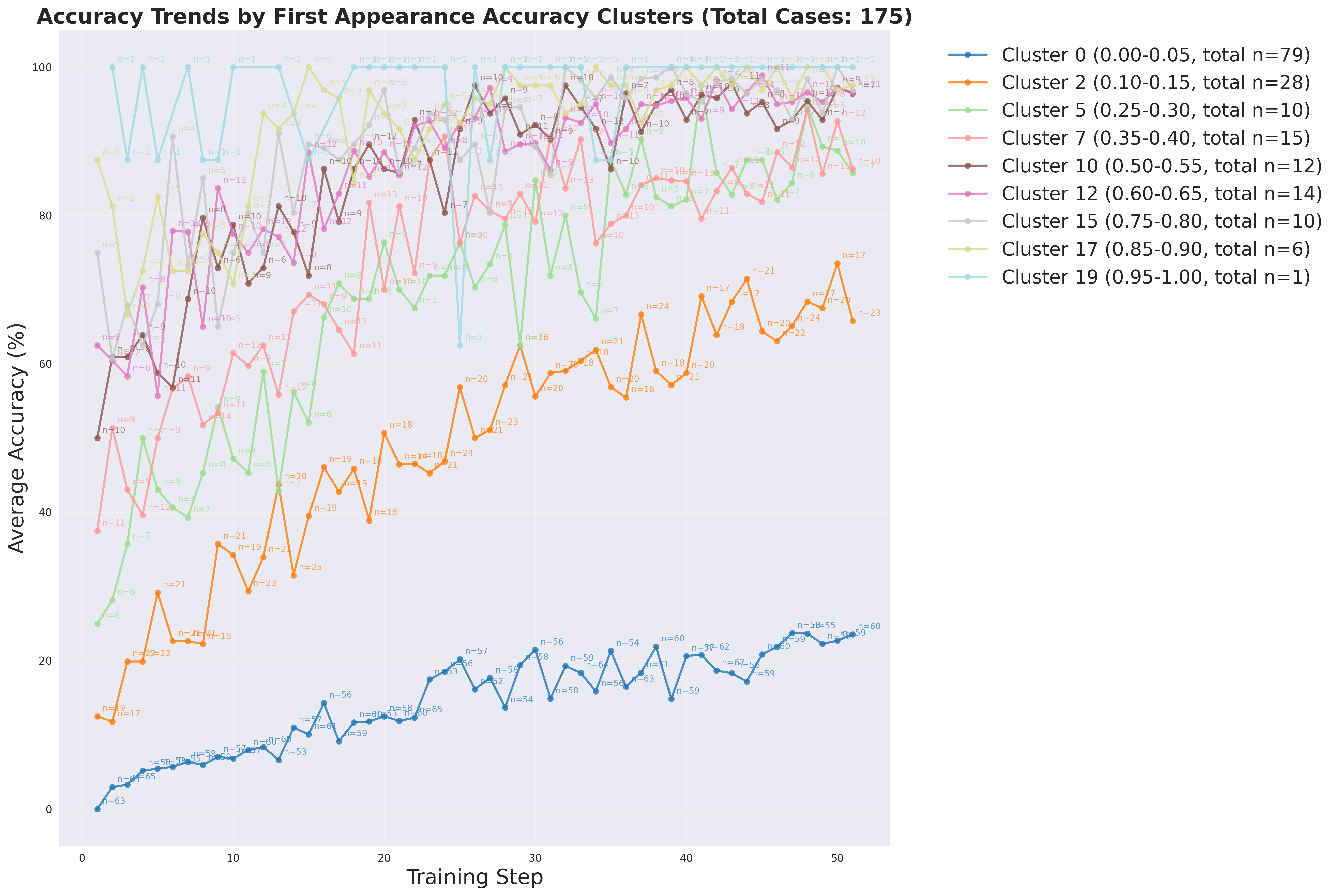}
    \caption{The Accuracy Trends by First Apperance Accuracy Clusters}
    \label{fig:ACCCluster}
\end{figure}
\subsection{Further Analysis}
\textbf{Analysis of training dynamics during standard RL training on 175 LeetCodeV6 cases.} We investigate the training dynamics by clustering cases based on their initial rollout accuracy (using rollout = 8 as the evaluation metric) and monitoring the evolution of training accuracy across different difficulty clusters. As illustrated in Figure~\ref{fig:ACCCluster}, cases with medium initial accuracy demonstrate the most rapid improvement, substantially outperforming both low- and high-accuracy clusters. While the modest gains in high-accuracy clusters are expected—these problems are already largely solved—the stagnant progress in low-accuracy clusters raises concerns about the model's ability to master challenging problems. This pattern indicates that standard RL training struggles with difficult cases, potentially creating a capability ceiling that limits the model's performance on complex problem-solving tasks. These findings motivate our two-stage approach, which specifically addresses this limitation through targeted curriculum learning.

\begin{figure}[!ht]
    \centering
    \includegraphics[width=0.85\textwidth]{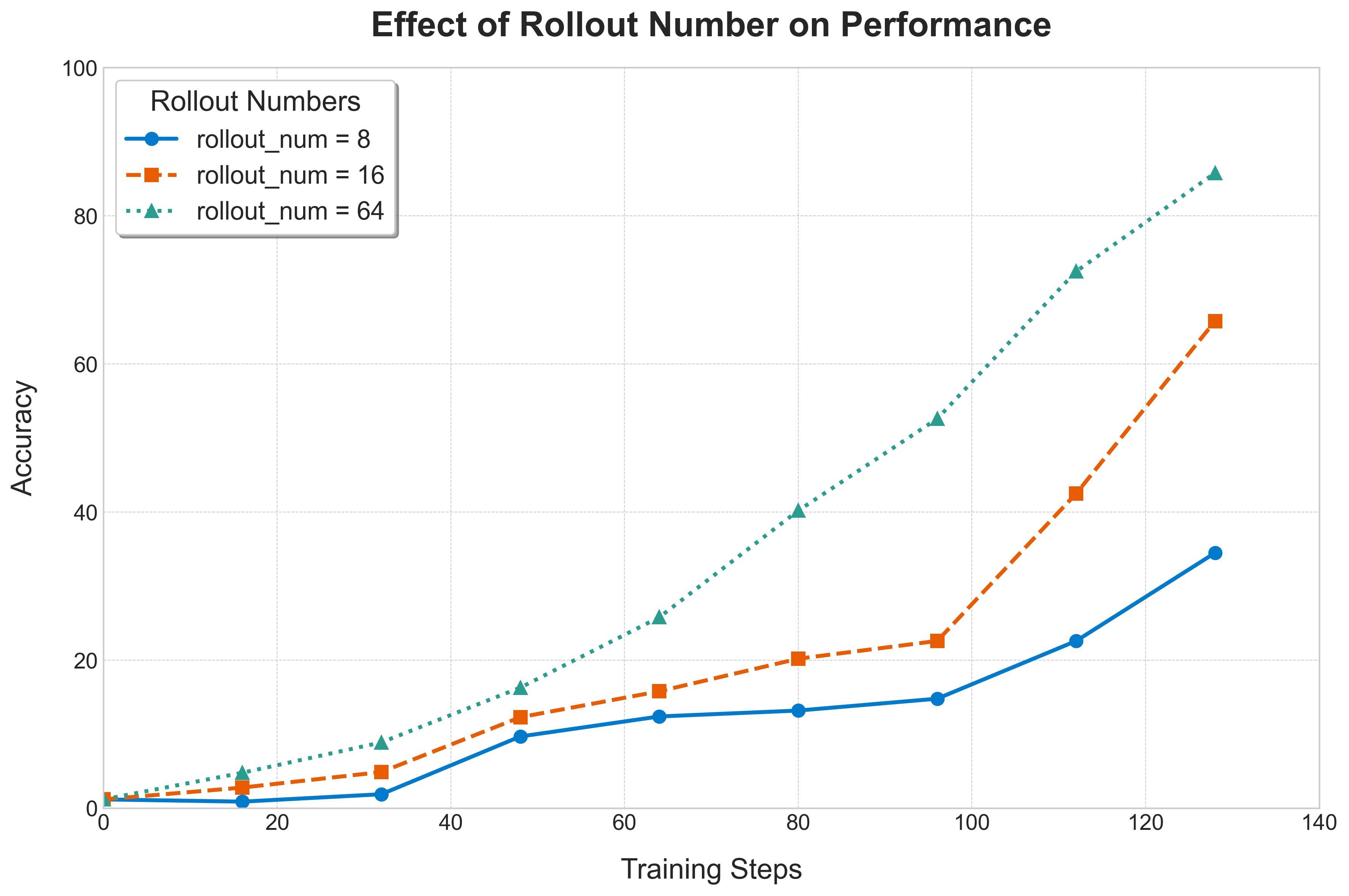}
    \caption{Effect of Rollout Number on Single-Case Training Performance}
    \label{fig:LargeRollout}
\end{figure}

\textbf{Analysis of few-case and single-case training.}
To obtain a finer-grained view of how the model learns individual cases, we trained on drastically reduced datasets—four cases and, in the extreme, a single case. The results mirror those from large-scale training: certain “hard” cases are exceptionally difficult to master. Even when the model was trained exclusively on a single hard case for 60 steps, it struggled to reach satisfactory performance. This motivates our second-stage strategy, which combines \textbf{large rollouts} with targeted focus on hard cases. The goal is to compel the model to learn these challenging examples, thereby pushing its capabilities and extending its generalization frontier. More details are provided in Appendix \ref{app:FewCaseTrain}.

\textbf{Ablation of the large-rollout setting in single-case training.}
Due to computational constraints, we conduct this ablation in the single-case training setting. Consistent with the analysis above, some hard cases remain difficult to learn even in this setting. Intuitively, increasing the rollout count should improve fitting efficiency. As shown in Figure \ref{fig:LargeRollout}, higher rollout counts accelerate learning. Furthermore, we observe: 1) Training on a single case has negligible impact on the performance of other cases, including easy ones. 2) Learning a hard case can improve generalization to other hard cases. More details are provided in Appendix \ref{app:FewCaseTrain}.

\textbf{More challenge training samples in the stage2.} Although harder samples in LiveCodeBench V6 already boost performance, we further investigated whether increasing the proportion of difficult training examples would continue to help. We curated 109 hard cases from our internal competitive-programming dataset (out of 633 candidates) and incorporated them into Stage 2. As shown in Table \ref{tab:ablation_study_wrapped}, training with a larger share of challenging samples consistently improves results across all benchmarks. Overall, this supports the conclusion that targeted inclusion of hard cases is an effective and scalable lever for improving model performance and generalization under limited compute.

\subsection{Scaling Trends}
To evaluate the scalability and effectiveness of our training methodology on a large-scale model, we applied it to competitive programming tasks for both SFT and RLVR. The experiments were conducted using a large-scale internal Mixture-of-Experts (MoE) model.

Our two-stage training process was configured as follows:
\begin{itemize}[left=0pt]
    \item \textbf{Stage 1 (Entropy Expansion):} The model was trained for 30 steps on a dataset of 9k samples. During this stage, we used a batch size of 512 with a rollout number of 16.
    
    \item \textbf{Stage 2 (Curriculum-based RL):} This stage consisted of a three-phase curriculum totaling 50 steps: 20 steps for Phase 1, 20 for Phase 2, and 10 for Phase 3. For this RL phase, the training configuration was a batch size of 128 with a larger rollout number of 64.
\end{itemize}

The performance results of this process are detailed in Table~\ref{tab:scaletrends}. An analysis of the table reveals several key insights:

\begin{itemize}[left=0pt]

\item \textbf{Baseline Performance:} The initial SFT model serves as a strong baseline, establishing the model's capabilities before RL fine-tuning.

\item \textbf{Intermediate Stage Effect:} After the entropy-expansion first stage, the model shows improved performance on some internal benchmarks (e.g., LiveCodeBench-V5) but a slight regression on external, more challenging ones like LeetCode and Codeforces weakly OJ. This suggests that while exploration broadens the model's policy, it requires the targeted learning of Stage 2 to become effective.

\item \textbf{Final Stage Efficacy:} Upon completing the second stage, the model demonstrates substantial and consistent performance gains across all benchmarks. As highlighted by the \textbf{Relative Improvement} metrics, our final model achieves gains of up to \textbf{+15.17\%} on the LeetCode Weekly OJ and \textbf{+7.51\%} on LiveCodeBench-V5 compared to the SFT baseline.
\end{itemize}

These results validate the effectiveness of our two-stage strategy and demonstrate positive scaling trends when applying RL with a carefully designed curriculum. It is important to note that these experiments were constrained by computational resources, and training was not continued to full convergence. For future work, we plan to integrate a greater number of challenging problems into the second stage curriculum. We anticipate this will lead to further performance gains, which will be presented in a subsequent formal report.

\begin{table}[t]
\centering
\caption{Scaling trends for RL training strategies. Metrics are decimals (3 d.p.); relative improvements are percentages.}
\label{tab:scaletrends}
\setlength{\tabcolsep}{6pt}
\renewcommand{\arraystretch}{1.2}
\begin{tabular}{lccccc}
\toprule
\textbf{Method} &
\shortstack{\textbf{LiveCode}\\\textbf{08--11}} &
\shortstack{\textbf{LiveCode}\\\textbf{V5}} &
\shortstack{\textbf{LiveCode}\\\textbf{V6}} &
\shortstack{\textbf{LeetCode Weekly (32)}\\\textbf{avg@1}} &
\shortstack{\textbf{Codeforces (33)}\\\textbf{avg@1}} \\
\midrule
\rowcolor{gray!10}
SFT model (64k) & 0.681 & 0.692 & 0.656 & 0.627 & 0.155 \\
\rowcolor{blue!10}
\shortstack[l]{RL Stage 1\\(32k train, 64k test)} & 0.690 & 0.740 & 0.665 & 0.611 & 0.123 \\
\rowcolor{green!20}
\shortstack[l]{RL Stage 2\\(64k, train step=50)} & 0.708 & 0.744 & 0.737 & 0.722 & 0.194 \\
\midrule
\rowcolor{yellow!12}
\shortstack{\textbf{Relative Improvement}\\\textbf{(Stage 1 vs SFT)}} & \textbf{+1.32\%} & \textbf{+6.94\%} & \textbf{+1.37\%} & \textbf{-2.55\%} & \textbf{-20.65\%} \\
\rowcolor{yellow!12}
\shortstack{\textbf{Relative Improvement}\\\textbf{(Stage 2 vs SFT)}} & \textbf{+3.96\%} & \textbf{+7.51\%} & \textbf{+12.35\%} & \textbf{+15.17\%} & \textbf{+25.16\%} \\
\bottomrule
\end{tabular}
\end{table}

\section{Conclusion}

In this work, we presented a two-stage reinforcement learning framework for competitive programming that addresses key limitations of supervised fine-tuning through entropy expansion and hard-focus curriculum learning. Our approach achieves state-of-the-art performance among 32B parameter models, with improvements ranging from 13\% to 58\% across various benchmarks, demonstrating that careful data curation and curriculum design are as crucial as algorithmic innovations for RLVR success.

Our analysis reveals three critical insights: (1) standard RL training struggles with hard problems, creating a capability ceiling that limits model performance; (2) large rollout budgets (64+ samples) are essential for mastering challenging cases; and (3) progressive curriculum learning that continuously retains the hardest problems outperforms uniform difficulty distribution. Ablation studies confirm both stages are necessary---entropy expansion enables robust generalization while hard-focus curriculum pushes the problem-solving frontier.

The scaling experiments on an internal large-scale MoE model validate that our principles transfer to larger scales. While our approach requires substantial computational resources for large rollout sampling, it provides a practical roadmap for training models capable of tackling complex algorithmic challenges. Future work could explore adaptive curriculum strategies and more efficient sampling techniques to further improve the cost-effectiveness of RLVR training for competitive programming.

\bibliography{LIMARLVR}
\bibliographystyle{LIMARLVR}

\newpage
\appendix
\onecolumn

\section{Few Cases Training}
\label{app:FewCaseTrain}
\begin{figure}[htbp]
\centering
\begin{subfigure}[b]{0.48\textwidth}
    \centering
    \includegraphics[width=\linewidth]{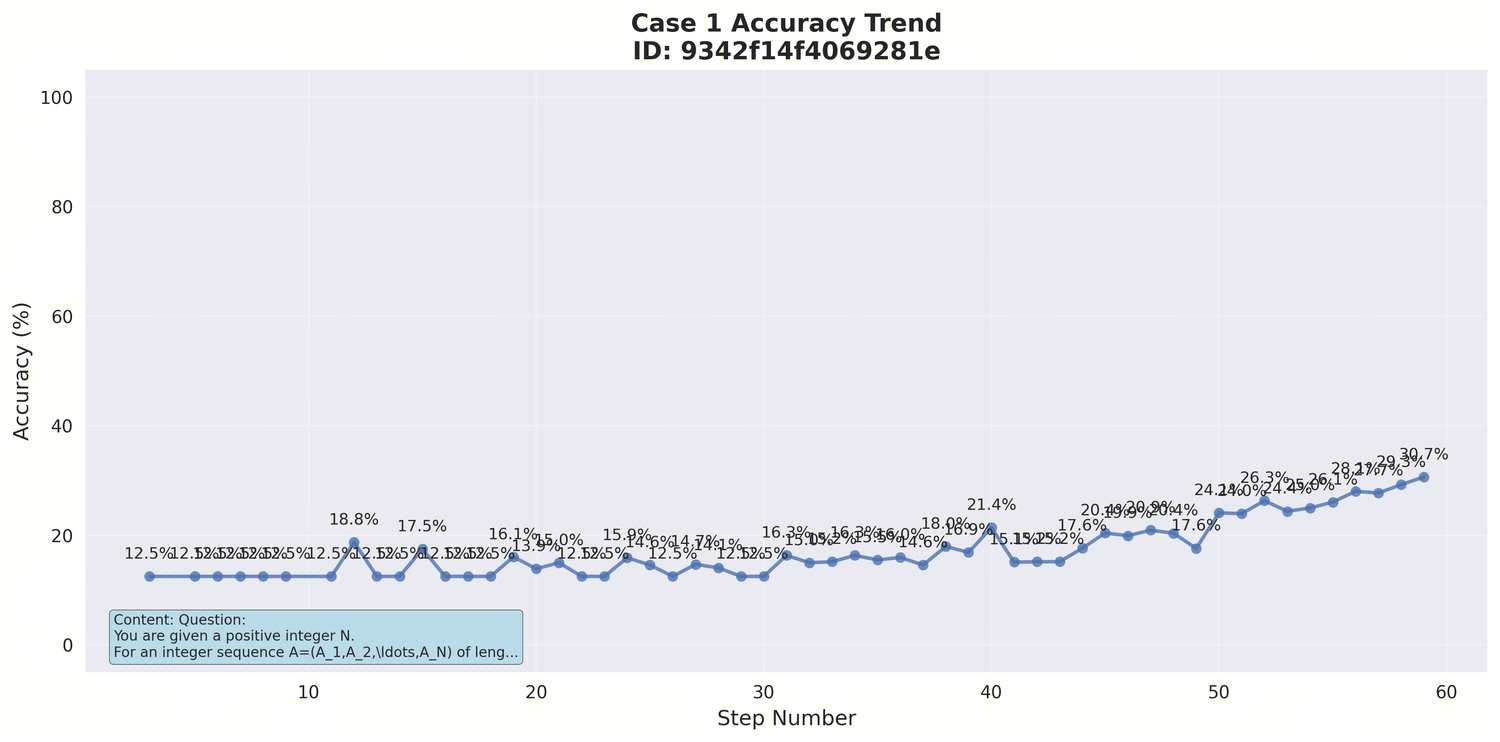}
    \caption{Case 1 Accuracy Evolution During Few-Shot Training}
    \label{fig:sub1}
\end{subfigure}
\hfill
\begin{subfigure}[b]{0.48\textwidth}
    \centering
    \includegraphics[width=\linewidth]{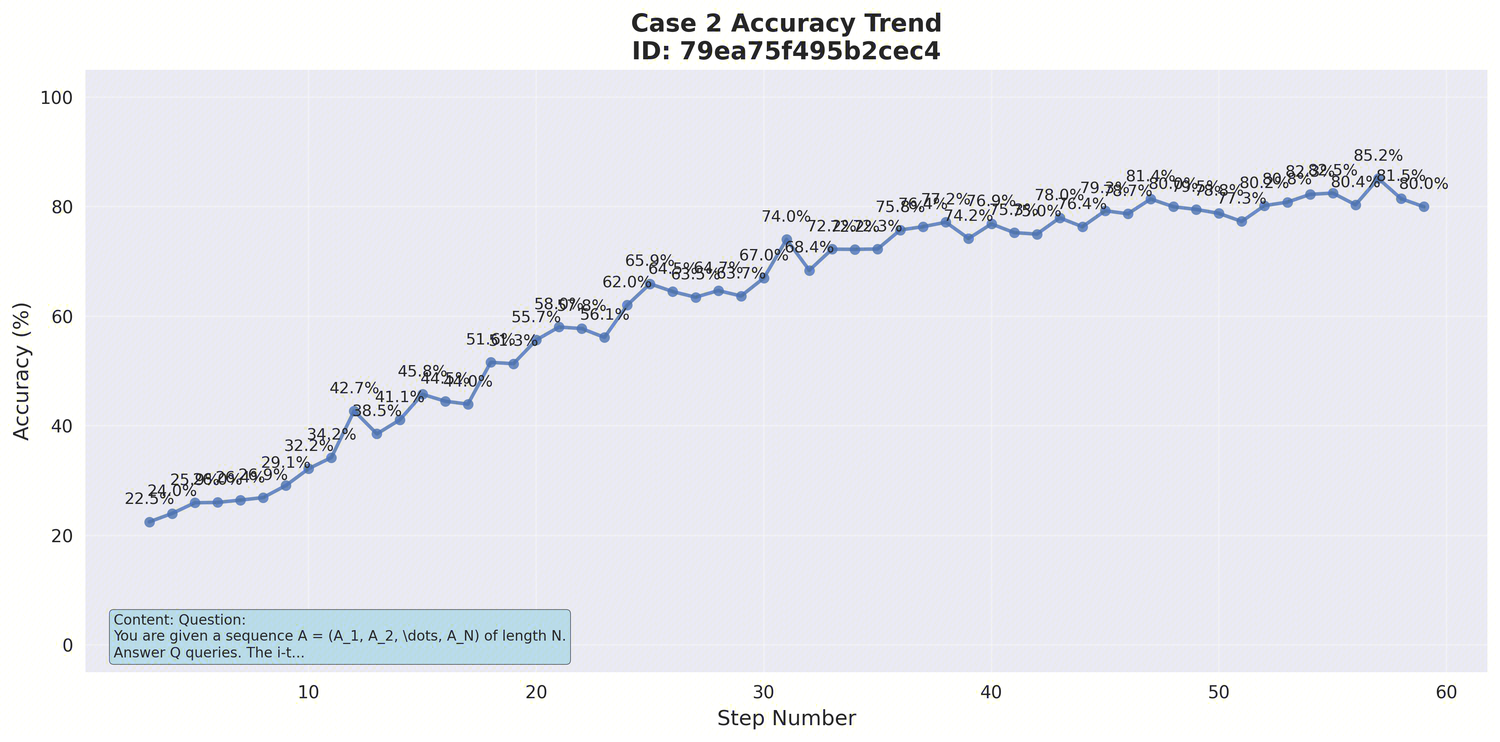}
    \caption{Case 2 Accuracy Evolution During Few-Shot Training}
    \label{fig:sub2}
\end{subfigure}
\\[1ex]
\begin{subfigure}[b]{0.48\textwidth}
    \centering
    \includegraphics[width=\linewidth]{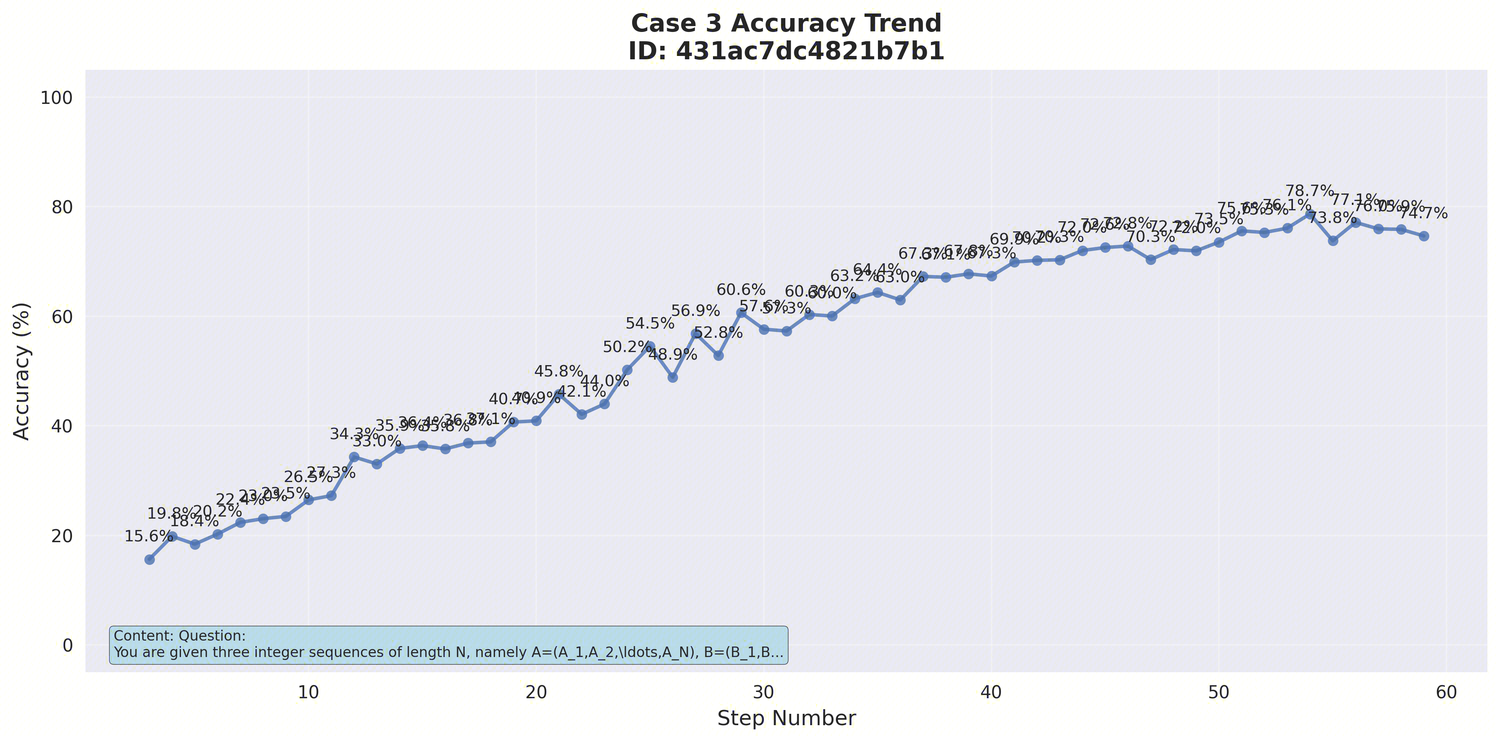}
    \caption{Case 3 Accuracy Evolution During Few-Shot Training}
    \label{fig:sub3}
\end{subfigure}
\hfill
\begin{subfigure}[b]{0.48\textwidth}
    \centering
    \includegraphics[width=\linewidth]{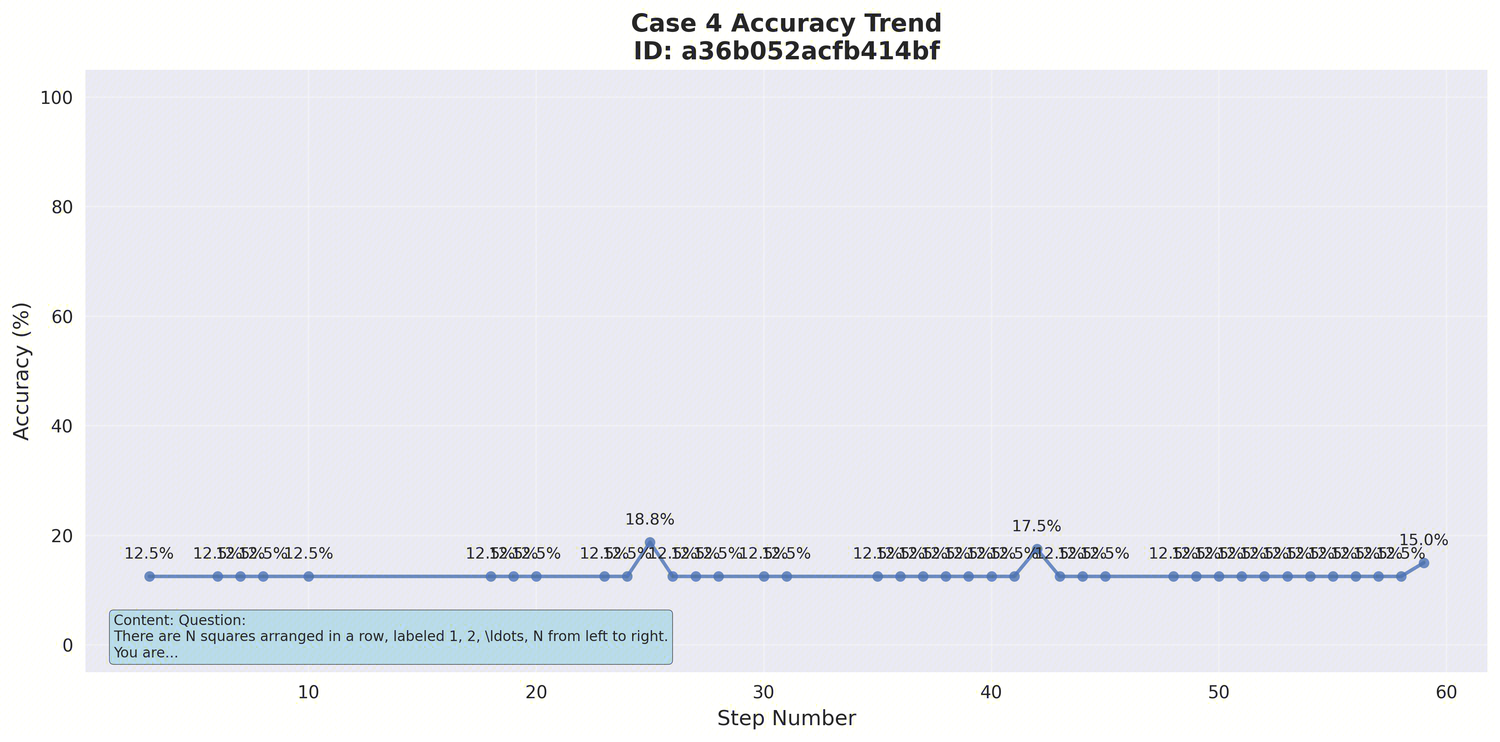}
    \caption{Case 4 Accuracy Evolution During Few-Shot Training}
    \label{fig:sub4}
\end{subfigure}
\caption{Case-wise Accuracy Trajectories Under Few-Shot Learning}
\label{fig:fourcases}
\end{figure}

\begin{figure}[!ht]
    \centering
    \includegraphics[width=0.8\textwidth]{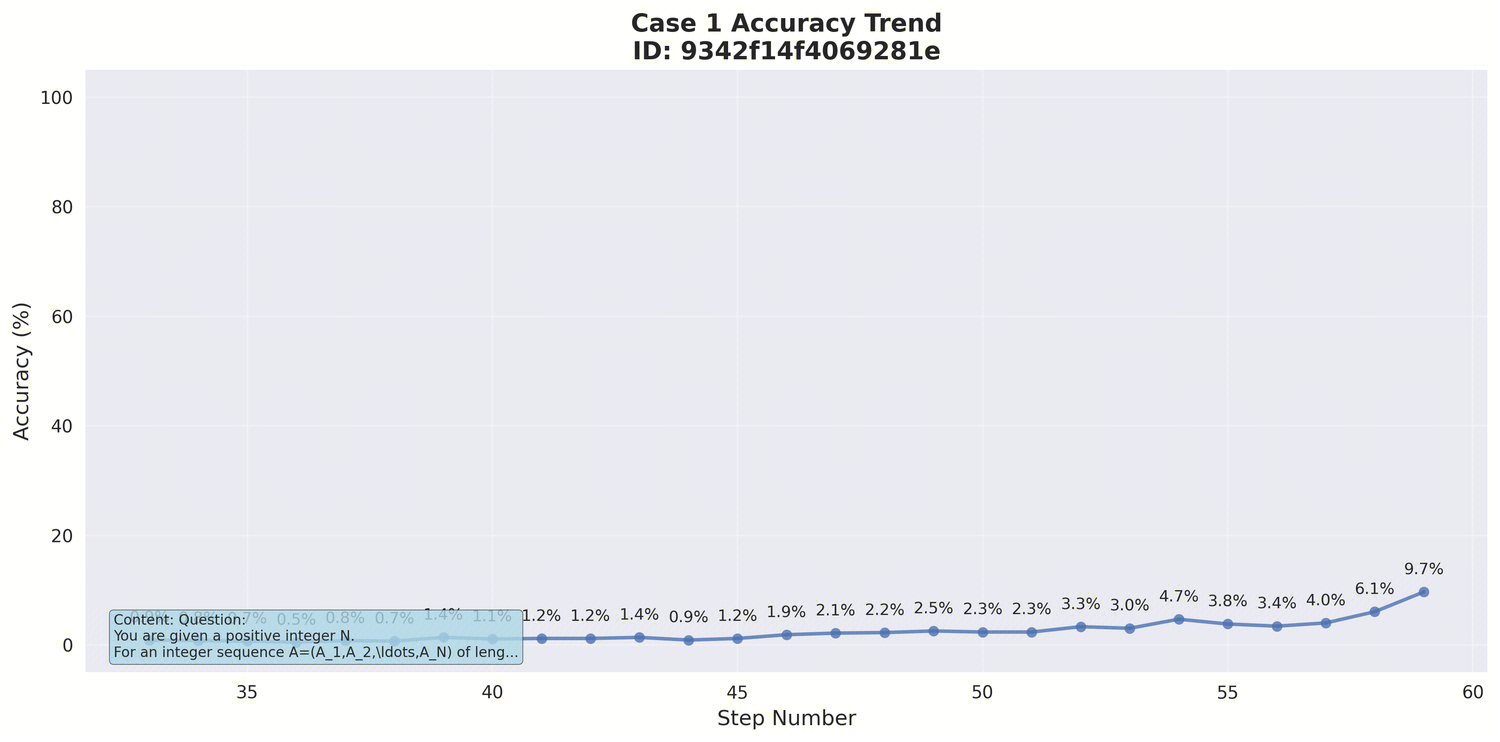}
    \caption{The Accuracy Trends by First Apperance Accuracy Clusters}
    \label{fig:onecases}
\end{figure}

To obtain a finer-grained view of how the model learns individual cases, we trained on drastically reduced datasets—four cases and, in the extreme, a single case. Specifically, as shown in Figure \ref{fig:fourcases}, we randomly selected 4 cases from LiveCodeBench V6's 175 problems and trained them using the GRPO algorithm with rollout num = 8. The learning curves reveal striking disparities in training dynamics across different problems. Cases 2 and 3 demonstrate rapid convergence, with accuracy surging from 15.6\% and 22\% to approximately 80\% within 60 steps, suggesting these problems align well with the model's inductive biases. In stark contrast, Case 1 exhibits only modest improvement from 12\% to 36.7\%, while Case 4 remains virtually frozen at 12.5\% throughout training, underscoring that certain "hard" cases pose exceptional learning challenges.

Further investigation through single-case training, as shown in Figure \ref{fig:onecases}, reveals an intriguing phenomenon: when Case 1 is trained in isolation, its learning trajectory becomes noticeably slower compared to the four-case setting. This observation provides evidence for positive transfer effects—the simultaneous training on Cases 2, 3, and 4 appears to facilitate Case 1's learning, suggesting that even diverse competitive programming problems share latent structural patterns that enable cross-problem generalization. However, the persistent difficulty with Case 1 (plateauing at merely 36.7\% after 60 steps) and the complete stagnation of Case 4 expose a fundamental training imbalance inherent in standard RL approaches, where the optimization process naturally gravitates toward easily learnable patterns while failing to break through on challenging problems. This phenomenon directly motivates our second-stage strategy employing large rollout budgets, specifically designed to compel the model to overcome these learning barriers on hard cases. 
\section{Case Study for Repetition Patterns}
\label{app:case}

When training models directly on algorithmic problem-solving tasks without extensive 24k-style training data (as demonstrated in the following case study), we observe significant repetition patterns that indicate inefficient reasoning strategies. This case study examines a graph isomorphism optimization problem where the model demonstrates various forms of computational redundancy and cyclical reasoning behaviors.

\begin{itemize}[left=0pt]
    \item \textbf{Computational Template Repetition}: The core permutation validation algorithm (inverse mapping construction, edge comparison, cost calculation) is applied systematically across dozens of different permutations, each following identical logical steps with different parameters.
    
    \item \textbf{Granular Analysis Loops}: Each permutation analysis involves repetitive edge-by-edge evaluation following the same pattern: determine desired state $\rightarrow$ check current state $\rightarrow$ calculate cost difference $\rightarrow$ accumulate total, creating highly formulaic micro-computations.
    
    \item \textbf{Hypothesis-Testing Cycles}: The reasoning exhibits recurring cycles of "try permutation X $\rightarrow$ calculate cost $\rightarrow$ compare to expected $\rightarrow$ express confusion $\rightarrow$ try next permutation," suggesting systematic but potentially inefficient exploration strategies.
    
    \item \textbf{Input Processing Redundancy}: Nearly identical parsing and data structure construction logic appears multiple times across different sample inputs, indicating missed abstraction opportunities in edge normalization and set construction.
    
    \item \textbf{Debugging Iteration Patterns}: Systematic testing of permutation variations (identity, single swaps, compound swaps) using identical verification procedures creates computational redundancy despite thorough coverage.
    
    \item \textbf{Meta-Cognitive Loops}: Higher-level reasoning patterns repeat the cycle of acknowledging complexity $\rightarrow$ attempting enumeration $\rightarrow$ encountering discrepancies $\rightarrow$ expressing uncertainty $\rightarrow$ restarting with variations, suggesting cognitive inefficiencies when facing algorithmic uncertainty.
\end{itemize}

These patterns suggest that without proper training on diverse reasoning examples within 24k context, models tend to fall into repetitive computational habits that, while methodical, lack the efficiency and abstraction capabilities demonstrated by more extensively trained systems.

Given the extensive length of the RL model's generated response in this case, we present only the analysis conducted by GPT-5, which identified numerous repetition patterns throughout the reasoning process.

\begin{lstlisting}[breaklines,numbers=left]
Looking at the content in `<think></think>`, I found extensive repetitive permutation calculation patterns. Here are the main repetitive patterns:

## Repetition Pattern 1: Basic Structure of Permutation Verification

```python
# This pattern repeats dozens of times in <think>
def verify_permutation_cost(p, n, G_set, H_set, cost):
    """Verify the cost calculation for permutation p"""
    q = [0] * n
    for i in range(n):
        q[p[i]] = i
    
    total_cost = 0
    for i in range(n):
        for j in range(i+1, n):
            u = min(q[i], q[j])
            v = max(q[i], q[j])
            desired = 1 if (u, v) in G_set else 0
            current = 1 if (i, j) in H_set else 0
            if desired != current:
                total_cost += cost[i][j]
    
    return total_cost
```

## Repetition Pattern 2: Edge State Checking Logic

```python
# Edge state checking that appears repeatedly in <think>
def check_edge_state(i, j, q, G_set, H_set):
    """Check the desired and current state of edge (i,j)"""
    u = min(q[i], q[j])
    v = max(q[i], q[j])
    desired = 1 if (u, v) in G_set else 0
    current = 1 if (i, j) in H_set else 0
    return desired, current
```

## Repetition Pattern 3: Detailed Permutation Calculations

There are many such repetitive calculations in `<think>`:

```python
# Example extraction of repetitive patterns
def detailed_permutation_analysis():
    """This detailed analysis repeats many times in think"""
    examples = [
        # p = [0, 1, 2, 3, 4] (identity)
        {
            "permutation": [0, 1, 2, 3, 4],
            "analysis": [
                "(0,1): desired=1, current=1 -> 0",
                "(0,2): desired=0, current=1 -> flip: cost[0][2]=1", 
                "(0,3): desired=0, current=1 -> flip: cost[0][3]=4",
                "(0,4): desired=0, current=1 -> flip: cost[0][4]=1",
                "(1,2): desired=1, current=0 -> flip: cost[1][2]=5",
                # ... more edge analyses
            ],
            "total": "1+4+1+5+6+3 = 20"
        },
        
        # p = [0, 1, 2, 4, 3] (swap 3 and 4)  
        {
            "permutation": [0, 1, 2, 4, 3],
            "analysis": [
                "(0,1): 0",
                "(0,2): desired=0, current=1 -> flip: cost[0][2]=1",
                "(0,3): desired=0, current=1 -> flip: cost[0][3]=4", 
                # ... similar repetitive analysis
            ],
            "total": "some calculation result"
        }
        # There are dozens of such examples in think
    ]
    return examples
```

## Repetition Pattern 4: Sample Verification Structure

```python
# Repetitive pattern for sample input processing
def process_sample_input(sample_name, n, G_edges, H_edges, cost_matrix, expected_output):
    """Common pattern for processing sample inputs"""
    print(f"Sample {sample_name}:")
    print(f"N={n}")
    print(f"G edges: {G_edges}")  
    print(f"H edges: {H_edges}")
    print(f"Cost matrix: {cost_matrix}")
    print(f"Expected: {expected_output}")
    
    # Convert to sets
    G_set = build_edge_set(G_edges)
    H_set = build_edge_set(H_edges)
    
    # Try various permutations
    min_cost = float('inf')
    for perm in all_permutations(n):
        cost = verify_permutation_cost(perm, n, G_set, H_set, cost_matrix)
        min_cost = min(min_cost, cost)
    
    print(f"Computed: {min_cost}")
    return min_cost == expected_output
```

## Repetition Pattern 5: Confusion Expression Statements

```python
# Confusion and retry statements that appear repeatedly in think
confusion_patterns = [
    "I am not finding 9.",
    "But the sample output is 9.", 
    "How do we get 9?",
    "This is not matching the sample operations.",
    "Given the complexity...",
    "How about we try the permutation: p = [...]",
    "total = ... -> Not X.",
    "But wait, ...",
    "I see the mistake: ...",
    "Therefore, ..."
]
```

These repetitive patterns indicate extensive trial-and-error processes and redundant calculations in the `<think>` section, which could be simplified by extracting common functions and establishing systematic verification workflows.
\end{lstlisting}

\end{document}